\documentclass{article}

\usepackage{microtype}
\usepackage{graphicx}
\usepackage{subcaption}
\usepackage{booktabs} % for professional tables
\usepackage[normalem]{ulem} 
\usepackage{hyperref}
\usepackage{multirow}

\usepackage[preprint]{icml2026}

\usepackage{amsmath}
\usepackage{amssymb}
\usepackage{mathtools}
\usepackage{amsthm}

\usepackage{url}
\usepackage{tcolorbox}
\usepackage{xcolor}
\usepackage{xspace}

\usepackage[capitalize,noabbrev]{cleveref}

\newcommand{\ourmodel}{\textsc{SwAP}\xspace}

\theoremstyle{plain}

\theoremstyle{definition}

\theoremstyle{remark}

\usepackage[textsize=tiny]{todonotes}
\newcommand{\icmlequalcontrib}{*Work done during internship at Amazon.}

\icmltitlerunning{Stepwise Penalization for Length-Efficient Chain-of-Thought Reasoning}

\begin{document}

\twocolumn[
  % \icmltitle{Mitigating Overthinking in Large Reasoning Models via Progress-Aware Stepwise Value Allocation}
  \icmltitle{Stepwise Penalization for Length-Efficient Chain-of-Thought Reasoning}



  \icmlsetsymbol{equal}{*}

  \begin{icmlauthorlist}
    \icmlauthor{Xintong Li}{equal,yyy}
    \icmlauthor{Sha Li}{comp}
    \icmlauthor{Rongmei Lin}{comp}
    \icmlauthor{Hongye Jin}{comp}
    \icmlauthor{Linwei Li}{comp}
    \icmlauthor{Hejie Cui}{comp}
    \icmlauthor{Sarah Zhang}{comp}
    \icmlauthor{Chia-Yuan Chang}{comp}
    \icmlauthor{Kewei Cheng}{comp}
    \icmlauthor{Besnik Fetahu}{comp}
    \icmlauthor{Priyanka Nigam}{comp}
    \icmlauthor{Jingbo Shang}{yyy}
    \icmlauthor{Bing Yin}{comp}
  \end{icmlauthorlist}

  \icmlaffiliation{yyy}{University of California, San Diego}
  \icmlaffiliation{comp}{Amazon}

  \icmlcorrespondingauthor{Xintong Li}{xil240@ucsd.edu}
  \icmlcorrespondingauthor{Sha Li}{slliz@amazon.com}
  \icmlcorrespondingauthor{Rongmei Lin}{linrongm@amazon.com}
  \icmlcorrespondingauthor{Hongye Jin}{hongyjin@amazon.com}

  \icmlkeywords{Machine Learning, ICML}

  \vskip 0.3in
]



\printAffiliationsAndNotice{\icmlequalcontrib}  % no special notice (required even if empty)

\begin{abstract}

% Large reasoning models achieve strong performance by scaling test-time computation. However, this paradigm often leads to overthinking, generating redundant chains-of-thought (CoT) that increase computational costs without improving accuracy.
% Existing reinforcement learning approaches typically address this by applying trajectory-level length penalties alongside a single outcome reward, which ignores step-wise heterogeneity, some steps are essential while others are redundant, and therefore tends to blunt compression. 
% To address this limitation, we propose a Progress-Aware Stepwise Reward framework, which enables fine-grained credit assignment to allocate length reduction efficiently across reasoning steps. 
% Our method estimates step importance intrinsically from the model’s on-policy log-probability improvements toward the correct answer. 
% Specifically, we quantify excess length as a penalty mass and redistribute it inversely proportional to step reward, thereby concentrating penalization on low-impact steps while preserving high-impact reasoning. 
% To optimize this objective under sparse verifiable outcomes, we combine the resulting stepwise signal with outcome rewards through a unified outcome–process advantage estimator within group-relative policy optimization, balancing global correctness with local efficiency.
% Experiments on mathematical reasoning benchmarks demonstrate that our approach substantially reduces generated reasoning length while maintaining strong task accuracy.

Large reasoning models improve with more test-time computation, but often overthink, producing unnecessarily long chains-of-thought that raise cost without improving accuracy. 
Prior reinforcement learning approaches typically rely on a single outcome reward with trajectory-level length penalties, which cannot distinguish essential from redundant reasoning steps and therefore yield blunt compression.
Although recent work incorporates step-level signals, such as offline pruning, supervised data construction, or verifier-based intermediate rewards, reasoning length is rarely treated as an explicit step-level optimization objective during RL.
We propose \textbf{S}tep-\textbf{w}ise \textbf{A}daptive \textbf{P}enalization (\ourmodel), a fine-grained framework that allocates length reduction across steps based on intrinsic contribution. 
We estimate step importance from the model's on-policy log-probability improvement toward the correct answer, then treat excess length as a penalty mass redistributed to penalize low-importance steps more heavily while preserving high-importance reasoning. 
We optimize with a unified outcome–process advantage within group-relative policy optimization. 
Experiments demonstrate that \ourmodel reduces reasoning length by \textbf{64.3\%} on average while improving accuracy by \textbf{5.7\%} relative to the base model.

\end{abstract}

% Step-Weighted Length Penalty
\section{Introduction}
Large language models (LLMs) have achieved strong performance in complex reasoning tasks through Chain-of-Thought (CoT) prompting, which encourages the model to decompose a problem into intermediate reasoning steps~\cite{achiam2023gpt, guo2025deepseek}. 
While this enables models to navigate complex logical structures, it often leads to overthinking: models generate excessively long reasoning chains dominated by redundant or low-value steps without improving final accuracy~\cite{sui2025stop}.
Such unnecessary verbosity substantially increases inference cost and latency, and can even degrade performance by introducing opportunities for hallucination in later stages of generation~\cite{li2025compressing, liu2025answer}.

\begin{figure}[t]
    \centering
    \includegraphics[
        width=\linewidth,
        trim=0.6cm 0.4cm 0.5cm 0.4cm,
        clip
    ]{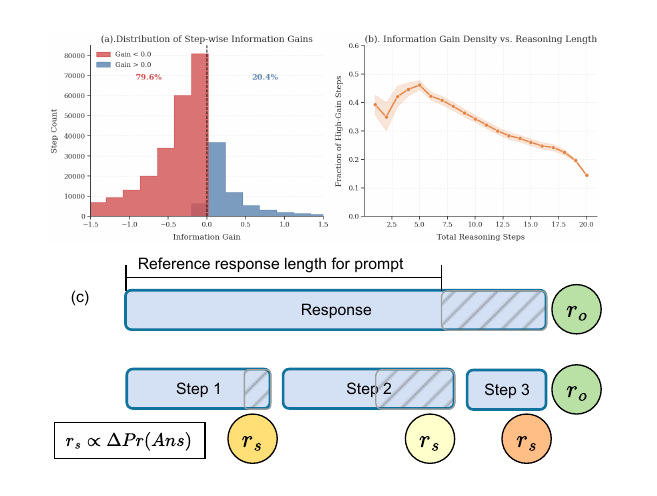}
    \caption{\textbf{Step-level redundancy in reasoning.}
    We segment the model's responses into steps by token length and use the change in the probability of generating the ground truth answer to measure each step's progress.
    (a) Distribution of step-wise information gain: most steps contribute little or no progress toward the correct answer, and high-gain steps are rare (4\% with gain $>$ 0.5).
    (b) Longer reasoning trajectories contain a lower fraction of high-gain steps.
    (c) In \ourmodel, in addition to the outcome reward $r_o$, each step also gets assigned a progress reward $r_s$. When the response is correct but exceeds the reference length, the penalty (dashed grey area) gets distributed to each step based on its progress (darker shades represent more progress).}
    \label{fig:step_analysis}
    \vspace{-4mm}
\end{figure}

To mitigate overthinking, recent work has increasingly turned to post-training strategies, including supervised preference learning for shorter responses~\cite{muennighoff2025s1}, distillation from long to short CoT models~\cite{li2025naturalthoughts}, and reinforcement learning (RL) with explicit length constraints~\cite{li2025aalc}.
Among these, RL-based approaches offer a particularly flexible mechanism to trade off accuracy and efficiency.
Most existing methods, however, operate at the trajectory level, for example, by imposing global length penalties, fixed token budgets, or difficulty-dependent reasoning modes~\cite{aggarwal2025l1,lyu2025hierarchical}.
Although effective in reducing the overall length of the reasoning, these coarse-grained strategies suffer from a fundamental limitation by treating all the reasoning steps equally valuable.

In practice, reasoning trajectories are highly heterogeneous at the step level~\cite{bogdan2025thought}.
Some steps introduce critical information that decisively pivots the solution toward correctness, while others correspond to redundant verification, reformulation, or low-utility exploration.
As a result, trajectory-level penalties often induce indiscriminate compression, risking the removal of essential reasoning while preserving irrelevant text~\cite{yu2025explainable}.
Figure~\ref{fig:step_analysis} empirically demonstrates this imbalance. As shown in (a), the majority of reasoning steps yield negligible or no information gain toward the final answer, with only a small fraction contributing substantial progress. 
Moreover, (b) shows that longer reasoning trajectories contain a markedly lower density of high-impact steps, indicating that redundancy accumulates as reasoning length increases.
Together, these observations indicate that overthinking is not simply a function of how much reasoning is performed, but of where redundant computation accumulates within a trajectory.

These findings highlight two principles that are largely underexplored in existing reward design.
First, length regularization should be aware of the difficulty. Harder problems naturally require longer reasoning and should not be penalized identically to easier ones~\cite{xiang2025just, song2025walk}.
Second, and more critically, length control should be conduct at the step level, rather than applied uniformly across an entire trajectory within a single trajectory, different reasoning steps contribute unequally to correctness and should therefore be treated asymmetrically~\cite{song2025mitigating}.
While recent work has begun to incorporate step-level information, such as offline pruning, supervised data construction, or verifier-based intermediate rewards, most approaches either use step signals outside of RL or rely on auxiliary models to assess step correctness~\cite{yue2508promoting, he2025smartthinker, wu2025ctrls}.
As a result, reasoning length itself is rarely treated as an explicit step-level optimization objective during RL, and existing methods lack a principled mechanism for deciding which steps to shorten and by how much when a trajectory becomes overly long.

In this work, we propose \textbf{S}tep-\textbf{w}ise \textbf{A}daptive \textbf{P}enalization (\ourmodel), a fine-grained RL framework that allocates length reduction across reasoning steps based on their intrinsic contribution to solving the problem.
Rather than relying on external reward models or heuristic estimators, \ourmodel derives step importance directly from the model’s own behavior.
Specifically, we quantify the information gain of each reasoning step by measuring the improvement in the log-probability assigned to the correct answer after that step.
Steps that induce large gains are interpreted as carrying substantial reasoning mass and should be preserved, while steps with minimal or no gain are treated as low-impact and preferentially penalized.

To preserve correctness while enforcing efficiency, we implement this algorithm via a stepwise weighted length penalty mechanism.
For each prompt, we define a reference length that is based on the difficulty of the correct responses and compute a global penalty when a trajectory exceeds this target.
This penalty mass is then redistributed across steps according to their importance, so that low-utility steps absorb most of the penalty while high-impact steps are protected from aggressive compression.
This design enables adaptive, fine-grained length control without requiring pre-specified token budgets or external supervision.
To ensure correctness while enforcing efficiency, we integrate the stepwise signal with trajectory-level outcome rewards through a unified outcome–process advantage under Group Relative Policy Optimization (GRPO)~\cite{guo2025deepseek}.
The outcome advantage promotes globally correct solutions, while a backward-propagated process advantage assigns credit to tokens based on the cumulative contribution of future reasoning steps.
Together, this unified formulation decomposes learning into complementary roles: outcome rewards gate correctness, while progress-aware step rewards shape where and how reasoning is compressed.

Experiments on five mathematical reasoning benchmarks demonstrate that \ourmodel achieves a substantially better accuracy–efficiency trade-off than prior methods.
For example, on DeepSeek-Distill-Qwen-1.5B, \ourmodel reduces reasoning length by \textbf{64.3\%} on average while improving accuracy by \textbf{5.7\%} relative to the base model.
On the 7B model, \ourmodel reduces token usage by over \textbf{50\%} while consistently matching or exceeding the strongest baselines on the hardest benchmarks, including AIME24, AIME25, and OlympiadBench.
These results indicate that significant overthinking persists even in large reasoning models, and that stepwise optimization provides a principled and effective path toward efficient reasoning.

\section{Related Work}

\paragraph{Efficient reasoning in LLMs.}
Modern reasoning LLMs often suffer from overthinking, producing unnecessarily long CoT that increase latency and inference cost without improving accuracy~\cite{feng2025efficient, sui2025stop}. Recent work on efficient reasoning spans both supervised fine-tuning and preference-based post-training. For instance, O1-Pruner~\cite{luo2025o1} applies length-harmonized fine-tuning to prune redundant reasoning while preserving performance, while AdaptThink~\cite{zhang2025adaptthink} introduces hybrid strategies that dynamically switch between ``thinking'' and ``no-thinking'' modes based on problem difficulty. SABER~\cite{zhao2025saber} further enables user-controllable, token-budgeted reasoning at inference time. Complementary analyses highlight that inefficiency is not only a matter of how much reasoning is generated, but also where redundancy accumulates. RC-R1~\cite{cheng2025optimizing} explicitly targets such redundant segments while preserving a sufficient core of reasoning steps. In parallel, recent studies on reasoning distillation and trace selection show that data quality and selection criteria strongly shape distilled reasoning behavior~\cite{li2025naturalthoughts, chen2025unveiling}, underscoring the importance of fine-grained control over reasoning structure.

\paragraph{RL for reducing overthinking.}

A prominent line of work leverages RL to trade off accuracy and efficiency through explicit length constraints. Early approaches such as L1~\cite{aggarwal2025l1} demonstrate that RL can enforce strong length control under fixed token budgets. Subsequent methods introduce progressively tighter constraints or iterative pruning strategies to compress long CoT~\cite{hou2025thinkprune}, adaptive reward shaping and target-length policies~\cite{liu2025learn}, and accuracy–length control rules or hierarchical budgeting mechanisms~\cite{li2025aalc, lyu2025hierarchical}. While effective at reducing overall reasoning length, most of these approaches apply penalties uniformly across the trajectory, leading to indiscriminate compression.
More recent step-aware RL methods address this limitation by incorporating verifiable intermediate feedback~\cite{wang2025masked}. 
VSRM assigns rule-based verifiable rewards to encourage effective steps and penalize ineffective ones~\cite{yue2508promoting}, while SmartThinker combines supervised fine-tuning with step-wise RL to guide reasoning refinement~\cite{he2025smartthinker}. These advances highlight the importance of step-level credit assignment for suppressing overthinking.
% \section{Preliminaries}

% GRPO
\section{Framework}

We introduce a progress-aware stepwise reward framework that redistributes trajectory-level length penalties across reasoning steps based on their marginal contribution to reaching the correct answer, and optimizes this signal jointly with outcome correctness under GRPO.
Our approach consists of three stages: estimating intrinsic step importance from on-policy log-probability improvements, allocating length penalties to individual steps according to their importance, and integrating the resulting stepwise signal with the outcome reward through a unified outcome–process advantage estimator. 
The full algorithm is summarized in Algorithm~\ref{alg:pasr}.

\subsection{Step Reward Measurement}
\label{sec:step_reward}

To assign step rewards, we define a robust step segmentation scheme and introduce an intrinsic measure of step importance based on answer log-probability improvement.

\paragraph{Step segmentation by token budget.}
Unlike prior work that relies on heuristic sentence boundaries or explicit reasoning transition words (e.g., \emph{wait}, \emph{however}, \emph{therefore})~\cite{besta2025reasoning, wang2023guiding, xu2025softcot}, we define reasoning steps using a simple and robust token-based segmentation scheme.

Given a generated response, we extract the reasoning portion (e.g., content before \texttt{</think>}) and partition it into at most $K$ steps $\{s_k\}_{k=1}^K$ by grouping approximately a fixed token budget $M$ per step.
If the response would exceed $K$ steps, we adapt $M$ upward to ensure a bounded number of steps.
We additionally align step boundaries to the next new line to produce cleaner, sentence-level segments.
Under this construction, each step corresponds to a contiguous span of tokens and forms a semantically coherent chunk of reasoning.

\paragraph{Step reward via answer log-probability gain.}
We define step importance intrinsically using the model’s own confidence toward the correct answer.
Intuitively, a reasoning step is important if it substantially increases the likelihood of the final answer.

Let $x$ denote the prompt, $a^\star$ the ground-truth answer, and $\{s_1, \ldots, s_K\}$ the segmented reasoning steps.
For each prefix consisting of the first $k$ steps, we construct the conditional context
$x_k = (x, s_1, \ldots, s_k)$, and compute the average per-token log-probability of the correct answer,
\begin{equation*}
\ell_k = \frac{1}{|a^\star|} \log p_\theta\!\left(a^\star \mid x_k \right).
\end{equation*}
We also compute a baseline score $\ell_0$ using the prompt without any reasoning steps.
We then define a progress-aware step reward as monotone incremental information gain
\begin{equation}
\Delta_k = \max\Bigl(0,\; \ell_k - \max_{j<k} \ell_j \Bigr),
\qquad k = 1,\ldots,K .
\label{eq:monotone_gain}
\end{equation}

This construction attributes positive reward only when a step improves the model’s confidence beyond all previous steps, preventing later redundant steps from accumulating spurious credit.
Steps that do not introduce new information receive zero reward.
Each $\Delta_k$ is assigned to the end of the corresponding step in the token sequence, yielding sparse, step-aligned reward markers that are compatible with on-policy reinforcement learning.

\subsection{Step-Weighted Length Penalty Redistribution}
\label{sec:penalty}

To discourage overlong reasoning while preserving essential computation, we introduce a step-weighted length penalty that selectively penalizes low-importance reasoning steps when a trajectory exceeds an adaptive length target.

Under GRPO, responses are generated in groups corresponding to the same prompt.
For each group, we define a target length $L_{\text{target}}$ as the median length of \emph{correct} responses within the group, which serves as a difficulty-aware reference.
Given a response of length $L$, if $L \le L_{\text{target}}$, no length penalty is applied.
Otherwise, we compute a total penalty mass
\begin{equation}
P = \lambda \cdot \frac{L - L_{\text{target}}}{L_{\text{target}}},
\label{eq:penalty_mass}
\end{equation}
where $\lambda$ controls the overall strength of length regularization.
Rather than applying this penalty uniformly across tokens, we redistribute it according to step importance.
Let $g_k = \ell_k - \ell_{k-1}$ denote the local log-probability gain associated with step $k$.
We assign each step a penalty weight
\begin{equation*}
w_k \propto \exp(- g_k / \tau),
\label{eq:weights}
\end{equation*}
where $\tau$ is a temperature parameter.
This construction assigns larger weights to steps with smaller gains, while protecting high-impact steps from aggressive penalization.
The weights are normalized such that $\sum_{k=1}^{K} w_k = 1$, and each step $k$ receives a penalty of $-P \cdot w_k$.

Finally, we combine intrinsic progress reward and length penalty into a single step-level reward,
\begin{equation}
r_k = \Delta_k - P \cdot w_k ,
\label{eq:final_step_reward}
\end{equation}
where $\Delta_k$ is the monotone information gain defined in Eq.~\ref{eq:monotone_gain}.
This step-weighted redistribution mechanism ensures that excess length is preferentially removed from low-utility reasoning segments, enabling fine-grained and adaptive reasoning compression without indiscriminately shortening the entire trajectory.

\begin{table*}[t]
\centering
\small
\setlength{\tabcolsep}{6pt}
\resizebox{\linewidth}{!}{
\begin{tabular}{lccccccccccccc}
\toprule
\multirow{2}{*}{\textbf{Method}}
& \multicolumn{2}{c}{\textbf{MATH-500}} 
& \multicolumn{2}{c}{\textbf{AMC23}} 
& \multicolumn{2}{c}{\textbf{AIME24}} 
& \multicolumn{2}{c}{\textbf{AIME25}} 
& \multicolumn{2}{c}{\textbf{OlympiadBench}} 
& \multicolumn{2}{c}{\textbf{Avg.}} \\
\cmidrule(lr){2-3}
\cmidrule(lr){4-5}
\cmidrule(lr){6-7}
\cmidrule(lr){8-9}
\cmidrule(lr){10-11}
\cmidrule(lr){12-13}
& Acc$\uparrow$ & Len$\downarrow$
& Acc$\uparrow$ & Len$\downarrow$
& Acc$\uparrow$ & Len$\downarrow$
& Acc$\uparrow$ & Len$\downarrow$
& Acc$\uparrow$ & Len$\downarrow$
& Acc$\uparrow$ & Len$\downarrow$
\\
\midrule

\multicolumn{10}{l}{\textbf{DeepSeek-Distill-Qwen-1.5B}} \\
Original 
& 83.4 & 4777
& 70.0 & 7722
& 26.7 & 12571
& 23.5 & 12247
& 35.1 & 11557
& 47.7 & 9775
\\
ThinkPrune
& 83.1 & 2332
& 72.5 & 3791
& 24.7 & 7469
& 20.4 & 7024
& 37.8 & 5634
& 47.7 & 5250
\\
LC-R1
& 81.9 & 2249
& 68.2 & 3607
& 23.3 & 7061
& 18.7 & 6647
& 33.3 & 5199
& 45.1 & 4953
\\
AdaptThink
& 83.9 & 1863
& 70.0 & 3628
& 31.3 & 7236
& 22.3 & 7195
& 38.1 & 4900
& 49.1 & 4964
\\
Laser-D
& 85.2 & 2589
& 74.2 & 4227
& 30.8 & 7929
& 22.1 & 7387
& 41.6 & 6498
& 50.8 & 5726
\\
Laser-DE
& 85.1 & 2821
& 75.1 & 4503
& 31.0 & 8330
& 21.2 & 7954
& 42.0 & 6819
& 50.9 & 6085
\\
\textbf{\ourmodel (ours)}
& \textbf{86.5} & 1597
& \textbf{78.3} & 2645
& \textbf{33.9} & 4915
& \textbf{23.3} & 4627
& \textbf{45.1} & 3646
& \textbf{53.4} & 3486
\\
\midrule
\multicolumn{10}{l}{\textbf{DeepSeek-Distill-Qwen-7B}} \\
Original
& 93.0 & 3743
& 84.6 & 5794
& 52.3 & 10579
& 37.5 & 11076
& 53.6 & 10222
& 64.2 & 8283
\\
L1-Exact
& 91.9 & 2186
& 81.1 & 2589
& 43.5 & 4026
& 28.7 & 4042
& 49.6 & 3438
& 59.0 & 3256
\\
L1-Max
& 92.0 & 2134
& 79.5 & 2594
& 44.5 & 4139
& 31.4 & 4093
& 47.5 & 3392
& 59.0 & 3270
\\
LC-R1
& 89.9 & 1568
& 81.6 & 2922
& 49.5 & 6972
& 36.2 & 7657
& 52.4 & 5678
& 61.9 & 4959
\\
AdaptThink
& 91.8 & 1825
& 80.0 & 3835
& 52.9 & 8600
& 37.5 & 9548
& 54.5 & 7856
& 63.3 & 6333
\\
Laser-D
& 93.5 & 1961
& \textbf{87.0} & 2925
& 55.6 & 6108
& 37.9 & 6629
& 59.7 & 5244
& 66.7 & 4573
\\
Laser-DE
& \textbf{93.6} & 1898
& 85.0 & 2901
& 53.5 & 5968
& 37.9 & 6161
& 58.3 & 5111
& 65.7 & 4408
\\
\textbf{\ourmodel (ours)}
& \textbf{93.6} & 1573
& 85.6 & 2685
& \textbf{56.5} & 5744
& \textbf{38.1} & 5598
& \textbf{60.3} & 4760
& \textbf{66.8} & 4072
\\

\bottomrule
\end{tabular}
}
\caption{Performance and reasoning length on mathematical reasoning benchmarks. Avg. denotes the average pass@1 across all datasets.}
\vspace{-6mm}
\label{tab:main_results}
\end{table*}

\subsection{Unified Outcome--Process Advantage}
\label{sec:advantage}

We integrate stepwise rewards into GRPO by constructing a unified token-level advantage that combines standard group-relative outcome supervision with a backward-propagated process signal derived from step rewards.
This formulation preserves the stability of outcome-based GRPO while enabling fine-grained efficiency optimization at the reasoning-step level.

\paragraph{Group-relative outcome advantage.}
For each query $x$, we samples a group of $N$ responses $\{y_i\}_{i=1}^N$.
Each response receives a sparse outcome reward $r^{\text{out}}_i \in \mathbb{R}$, defined as positive if the final answer is correct and zero otherwise.
Following the standard GRPO, outcome rewards are normalized within the group using mean $\mu_{\text{out}}(x)$ and standard deviation $\sigma_{\text{out}}(x)$, resulting in the outcome advantage
\begin{equation}
A^{\text{out}}_i
=
\frac{r^{\text{out}}_i - \mu_{\text{out}}(x)}{\sigma_{\text{out}}(x) + \varepsilon}.
\label{eq:out_adv}
\end{equation}
% This scalar advantage is broadcast to all tokens in response $y_i$ during optimization.

\paragraph{Step-level rewards.}
Each response $y_i$ is segmented into $K_i$ reasoning steps, with step rewards $\{r_{i,k}\}_{k=1}^{K_i}$ defined in Eq.~\eqref{eq:final_step_reward} and placed at the end token of each step.
To stabilize training, step rewards are normalized within each group using statistics computed over \emph{correct} trajectories only,
\begin{equation}
\tilde r_{i,k}
=
\frac{r_{i,k} - \mu_{\text{step}}(x)}{\sigma_{\text{step}}(x) + \varepsilon},
\label{eq:norm_step}
\end{equation}
where $\mu_{\text{step}}(x)$ and $\sigma_{\text{step}}(x)$ denote the mean and standard deviation of step rewards across correct responses for the prompt $x$.

\paragraph{Backward-propagated process advantage.}
Let $e_{i,k}$ denote the token index corresponding to the end of step $k$ in response $y_i$.
We define a token-level process advantage by accumulating normalized step rewards from future reasoning steps,
\begin{equation}
A^{\text{proc}}_{i,t}
=
\sum_{k:\; t \le e_{i,k}} \tilde r_{i,k}.
\label{eq:proc_adv}
\end{equation}
This reverse cumulative construction assigns each token credit proportional to the cumulative contribution of future reasoning steps, allowing step-level efficiency signals to influence earlier generation decisions.

\begin{algorithm}[t]
\caption{\ourmodel}
\label{alg:pasr}
\small
\setlength{\abovedisplayskip}{2pt}
\setlength{\belowdisplayskip}{2pt}
\begin{algorithmic}[1]
\REQUIRE Policy $\pi_{\theta}$, dataset $\mathcal{D}$, group size $N$, weights $(\beta,\theta)$
\FOR{$t=1,\ldots,T$}
  \STATE Sample prompts $x \sim \mathcal{D}$ and rollouts $\{y_i\}_{i=1}^N \sim \pi_{\theta_{\mathrm{old}}}(\cdot \mid x)$
  \FOR{each rollout $y_i$}
    \STATE Segment into steps $\{s_{i,k}\}_{k=1}^{K_i}$ via token budget $M$
    \STATE Calculate step information gain $\Delta_{i,k}$ \textbf{(Eq.~\ref{eq:monotone_gain})}
    \STATE Penalty redistribution: compute penalty mass $P_i$ and weight $w_{i,k}\propto \exp(-g_{i,k}/\tau)$
    \STATE Step reward $r_{i,k}=\Delta_{i,k}-P_i w_{i,k}$ \textbf{(Eq.~\ref{eq:final_step_reward})}
  \ENDFOR
  \STATE Outcome advantage $A_i^{\mathrm{out}}=\dfrac{r_i^{\mathrm{out}}-\mu_{\mathrm{out}}(x)}{\sigma_{\mathrm{out}}(x)+\varepsilon}$
  \STATE Normalize step rewards on correct rollouts: $\tilde r_{i,k}$  \textbf{(Eq.~\ref{eq:norm_step})}
  \STATE Process advantage $A_{i,t}^{\mathrm{proc}}$ \textbf{(Eq.~\ref{eq:proc_adv})}
  \STATE Unified advantage $A_{i,t}=\beta A_i^{\mathrm{out}}+\theta\,\mathbb{I}[r_i^{\mathrm{out}}>0]A_{i,t}^{\mathrm{proc}}$
  \STATE Update $\theta$ with clipped GRPO objective using $A_{i,t}$ \textbf{(Eq.~\ref{eq:grpo_loss})}
\ENDFOR
\end{algorithmic}
\end{algorithm}

\paragraph{Unified advantage.}
Finally, we combine outcome and process advantages into a single token-level signal,
\begin{equation}
A_{i,t}
=
\beta \, A^{\text{out}}_i
\;+\;
\theta \, \mathbb{I}\!\left[r^{\text{out}}_i > 0\right] \cdot A^{\text{proc}}_{i,t},
\label{eq:unified_adv}
\end{equation}
where $\beta,\theta>0$ control the trade-off between global correctness and local efficiency.
Importantly, the process term is gated by correctness, ensuring that step-level signals influence optimization only for correct trajectories and preventing noisy step rewards from corrupting learning on incorrect samples.

Let $\pi_{\theta_{\text{old}}}$ denote the behavior policy used to generate rollouts.
We optimize the standard clipped GRPO objective using the unified advantage:
\begin{align}
\mathcal{L}_{\text{GRPO}}(\theta) = - &\mathbb{E}_{x,i,t} \Big[ \min \big(r_{i,t}(\theta) A_{i,t}, \nonumber \\
&\operatorname{clip}(r_{i,t}(\theta), 1-\epsilon, 1+\epsilon) A_{i,t} \big) \Big] \label{eq:grpo_loss}
\end{align}
where $r_{i,t}(\theta) = \frac{\pi_\theta\!\left(y_{i,t}\mid x,\, y_{i,<t}\right)}
{\pi_{\theta_{\text{old}}}\!\left(y_{i,t}\mid x,\, y_{i,<t}\right)} $ is the token-wise importance ratio.

Algorithm~\ref{alg:pasr} summarizes the overall SWAP training procedure, including step importance estimation, penalty redistribution, and unified advantage construction under GRPO.
\section{Experiments}
\label{sec:experiments}

\subsection{Experimental Setup}

\begin{table*}[t]
\centering
\small
\setlength{\tabcolsep}{6pt}
\resizebox{\linewidth}{!}{
\begin{tabular}{lccccccccccccc}
\toprule
\multirow{2}{*}{\textbf{Method}}
& \multicolumn{2}{c}{\textbf{MATH-500}} 
& \multicolumn{2}{c}{\textbf{AMC23}} 
& \multicolumn{2}{c}{\textbf{AIME24}} 
& \multicolumn{2}{c}{\textbf{AIME25}} 
& \multicolumn{2}{c}{\textbf{OlympiadBench}} 
& \multicolumn{2}{c}{\textbf{Avg.}} \\
\cmidrule(lr){2-3}
\cmidrule(lr){4-5}
\cmidrule(lr){6-7}
\cmidrule(lr){8-9}
\cmidrule(lr){10-11}
\cmidrule(lr){12-13}
& Acc$\uparrow$ & Len$\downarrow$
& Acc$\uparrow$ & Len$\downarrow$
& Acc$\uparrow$ & Len$\downarrow$
& Acc$\uparrow$ & Len$\downarrow$
& Acc$\uparrow$ & Len$\downarrow$
& Acc$\uparrow$ & Len$\downarrow$
\\
\midrule

Original
& 83.4 & 4777
& 70.0 & 7722
& 26.7 & 12571
& 23.5 & 12247
& 35.1 & 11557
& 47.7 & 9775 
\\
Outcome-Only
& 79.5 & 1534
& 70.6 & 2953
& 29.8 & 6287
& 21.8 & 6048
& 41.3 & 4433
& 48.6 & 4251  \\
Step-Only
& 79.9 & 1526
& 67.8 & 2915
& 27.1 & 6019
& 20.8 & 5872
& 31.5 & 3689
& 45.4 & 4004 \\
No-Penalty
& 85.7 & 2594
& 77.1 & 4169
& 32.5 & 8027
& 24.3 & 7602
& 42.7 & 6237
& 52.5 & 5726 \\
Static-Penalty
& 83.7 & 1276
& 72.8 & 1899
& 25.2 & 3264
& 20.6 & 2702
& 43.0 & 2110
& 49.1 & 2250 \\
Uniform-Penalty
& 86.4 & 2064
& 77.8 & 3469
& 32.9 & 6568
& \textbf{25.4} & 6163
& 44.5 & 5118
& \textbf{53.4} & 4676 \\
\textbf{\ourmodel (ours)}
& \textbf{86.5} & 1597
& \textbf{78.3} & 2645
& \textbf{33.9} & 4915
& 23.3 & 4627
& \textbf{45.1} & 3646
& \textbf{53.4} & 3486 \\
\bottomrule
\end{tabular}
}
\caption{Performance and reasoning length of different model variants. All results are based on the DeepSeek-Distill-Qwen-1.5B model. AVG denotes the average pass@1 across all datasets.}
\vspace{-6mm}
\label{tab:penalty}
\end{table*}

\paragraph{Models and Training Data.}
We evaluate \ourmodel on two widely used reasoning models, DeepSeek-R1-Distill-Qwen-1.5B and DeepSeek-R1-Distill-Qwen-7B~\cite{deepseekai2025deepseekr1incentivizingreasoningcapability}.
Both models demonstrate strong mathematical reasoning capability but frequently exhibit overthinking behavior, producing unnecessarily long chains-of-thought.
For training, we use the DeepScaleR-Preview dataset~\citep{luo2025deepscaler}, which contains 40K high-quality mathematical reasoning problems collected from AIME, AMC, Omni-Math~\cite{gao2024omni}, and STILL~\cite{min2024imitate}.
This dataset provides diverse difficulty levels and supports studying both reasoning accuracy and efficiency.

\paragraph{Baselines.}
We compare \ourmodel against a diverse set of state-of-the-art methods for efficient reasoning, covering three representative categories.
Length-penalty-based RL methods include ThinkPrune~\cite{hou2025thinkprune}, which progressively enforces stricter token limits during RL to prune redundant reasoning; LC-R1~\cite{cheng2025optimizing}, which combines GRPO with global length and compression rewards to remove invalid post-solution reasoning; and two variants of the LASER~\cite{liu2025learn} framework, LASER-D and LASER-DE, which apply dynamic and difficulty-aware length reward shaping, with LASER-DE additionally encouraging exploration on incorrect trajectories.
Adaptive thinking-mode methods include AdaptThink~\cite{zhang2025adaptthink}, which trains models to dynamically select between thinking and no-thinking modes based on problem difficulty.
Length-control methods include L1-Exact and L1-Max~\cite{aggarwal2025l1}, which optimize reasoning accuracy under prompt-specified exact or maximum length constraints via length-controlled policy optimization.

% All baseline methods are trained or evaluated under the same model architectures, datasets, and sampling settings as \ourmodel to ensure a fair comparison.

\paragraph{Evaluation Benchmarks.}
We evaluate all methods on five mathematical reasoning benchmarks spanning a wide range of difficulty levels:
MATH-500~\cite{hendrycks2021measuring}, AMC23~\cite{amc}, AIME24, AIME25~\cite{aime}, and OlympiadBench~\cite{he2024olympiadbench}.
Evaluation is performed with temperature $T=0.6$, top-$p=0.95$, and a maximum generation length of 32{,}768 tokens.
For each benchmark, we generate 16 samples per query and report both pass@1 accuracy and the average response length in tokens, capturing the trade-off between correctness and reasoning efficiency.

\paragraph{Implementation Details.}
We implement \ourmodel using the VeRL framework~\cite{sheng2025hybridflow}.
During training, we use a batch size of 128 and generate 16 rollouts per prompt with a maximum response length of 12{,}000 tokens and sampling temperature 0.6.
Training is conducted for two epochs for both model sizes, resulting in a total of 623 steps.

For step segmentation, we allocate approximately 350 tokens per step and cap the maximum number of reasoning steps at 25.
We set the length penalty coefficient $\lambda=1$ to apply the full penalty mass when responses exceed the target length.
The outcome advantage weight $\beta$ is set to 1, while the process advantage weight $\theta$ is set to 0.3 to avoid overly aggressive early length reduction.
Training of the 1.5B model is conducted on a single node with $8\times$H100 GPUs, whereas the 7B model is trained on a single node with $8\times$H200 GPUs.

\subsection{Main results}

\begin{figure*}[ht]
    \centering
    \includegraphics[width=\linewidth]{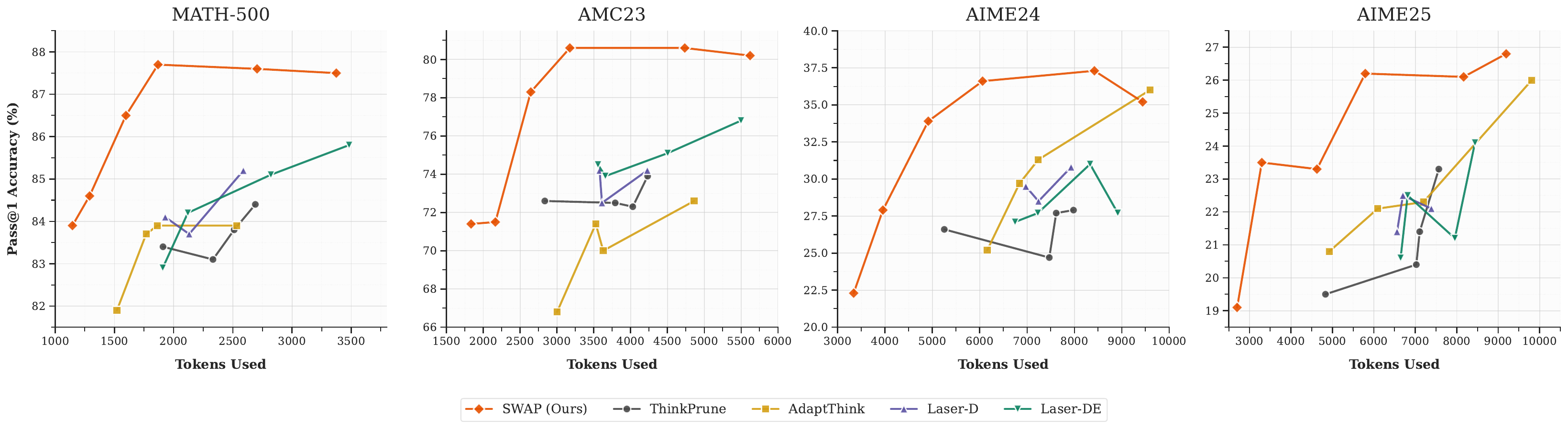}
    \caption{Comparison of Pass@1 accuracy vs. token usage across math benchmarks. \ourmodel establishes the Pareto frontier for best performance under every token budget.}
    \label{fig:dy_budget}
    \vspace{-2mm}
\end{figure*}
Table~\ref{tab:main_results} reports results on five mathematical reasoning benchmarks for both model scales. 
Across all settings, \ourmodel significantly reduces reasoning length while simultaneously improving accuracy, achieving the strongest accuracy–efficiency trade-off among all methods.

Relative to the base models, \ourmodel substantially compresses reasoning traces without sacrificing correctness. 
For the 1.5B model, \ourmodel reduces average response length by 64.3\% while improving average accuracy by 5.7\%, with consistent gains across all benchmarks and particularly large improvements on harder datasets such as AIME24 and OlympiadBench.
A similar trend holds for the 7B model, where \ourmodel reduces reasoning length by 50.8\% on average and improves accuracy on all benchmarks, achieving the strongest results on AIME24, AIME25, and OlympiadBench. 
These results suggest that substantial overthinking is present even among larger reasoning models, especially on difficult problems, and that progress-aware stepwise length reduction can effectively remove redundant computation.

Compared with efficient reasoning baselines, \ourmodel consistently achieves a more favorable accuracy–efficiency balance. 
Methods such as ThinkPrune and LC-R1 achieve notable length reductions but often degrade accuracy, especially on harder benchmarks, suggesting that trajectory-level penalties may indiscriminately remove essential reasoning steps alongside redundant ones. 
For the 7B model, L1-Exact and L1-Max enforce strict token budgets, reducing token usage by around 60\% but decreasing average accuracy by 5.2\%, with particularly large losses on challenging datasets.
Among all baselines, LASER-D and LASER-DE are the most competitive. However, \ourmodel consistently matches or exceeds their accuracy while using substantially fewer tokens. For example, on the 1.5B model, \ourmodel improves accuracy on AIME24 and OlympiadBench while reducing reasoning length by 35–45\% relative to LASER variants. Similar trends are observed for the 7B model, where \ourmodel achieves the best or tied-best accuracy on the hardest benchmarks while producing among the shortest reasoning traces overall.

\begin{figure}[ht]
    \centering
    \includegraphics[width=\linewidth]{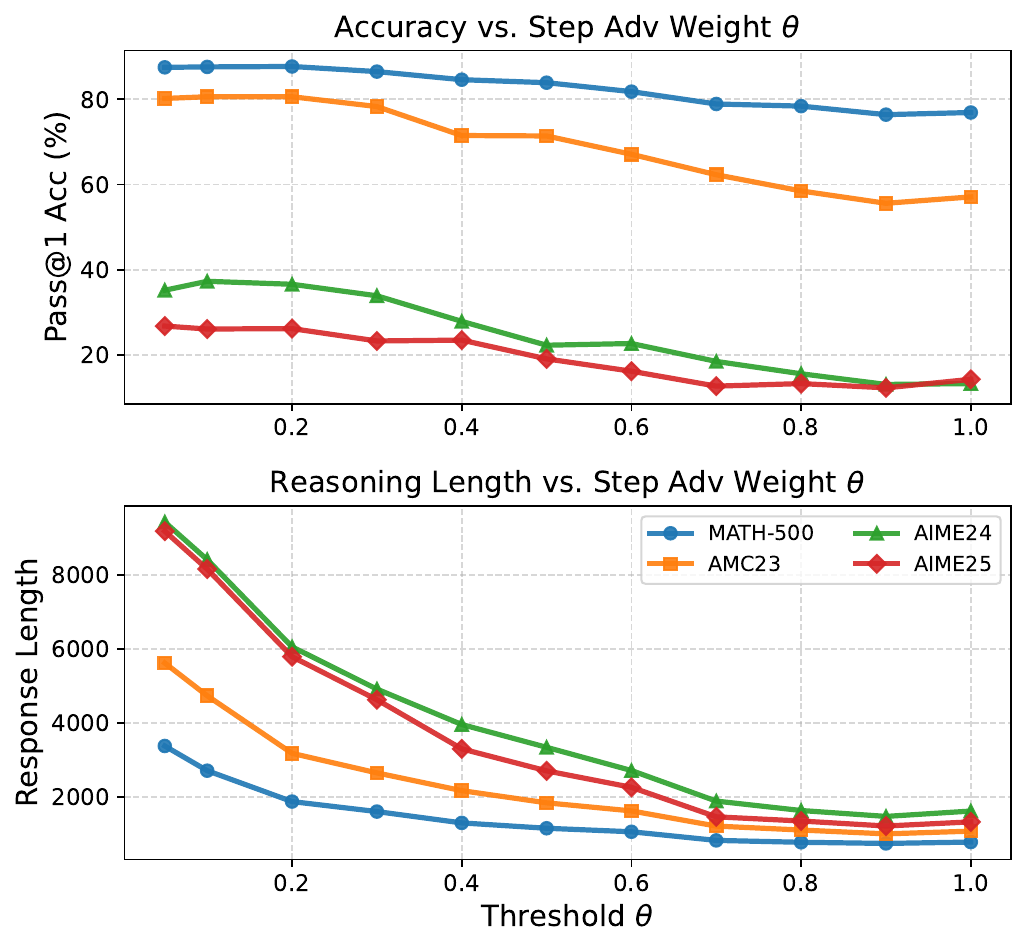}
    \caption{Impact of step advantage weight $\theta$ on model performance and response length. A moderate value of $\theta \in [0.2, 0.4]$ achieves the best balance. Hard datasets such as AIME 2024/2025 are more sensitive to a large $\theta$ value.}
    \label{fig:theta}
    \vspace{-6mm}
\end{figure}

\subsection{Reward Components and Penalty Strategies}

We study the contribution of each component in our framework using the DeepSeek-Distill-1.5B model. As shown in Table~\ref{tab:penalty}, we compare our unified approach against variants that isolate specific reward signals and penalty mechanisms.

\textbf{Necessity of Unified Advantages.} 
We first evaluate the performance of using either outcome or step signals in isolation. The \textbf{Outcome-Only} variant, which applies a trajectory-level length penalty to the final reward, successfully reduces reasoning length by over 56\% on average. 
However, it suffers from a significant drop in accuracy (e.g., -3.9\% on MATH-500), indicating that global signals induce indiscriminate compression that inadvertently removes steps essential for correctness.
Conversely, the \textbf{Step-Only} variant, relying solely on intrinsic progress, is the only method that performs worse than the baseline model (-2.3\%). 
This highlights that while stepwise rewards provide fine-grained signals, they lack the global grounding necessary to preserve overall logical integrity. 
These results confirm that the synergy between global outcome advantages and local process advantages is critical for balanced optimization.

\textbf{Length Penalty Design and Distribution}.
The absence of an explicit penalty (\textbf{No-Penalty}) results in the longest reasoning trajectories among all RL variants. 
While this setting achieves strong accuracy, its average length remains 64\% higher than \ourmodel, demonstrating that intrinsic step rewards alone are insufficient to suppress redundancy without explicit cost pressure.
\textbf{Static-Penalty} applies a uniform per-step penalty, while \textbf{Uniform-Penalty} distributes the total penalty mass evenly across steps.
Both reduce reasoning length more aggressively than No-Penalty; however, Static-Penalty severely degrades performance on challenging benchmarks such as AIME, where uniform discouragement likely suppresses the depth of reasoning required for complex problem solving.
Uniform-Penalty achieves better accuracy, matching \ourmodel’s average performance (53.4\%), but remains substantially less efficient. 
In comparison, \ourmodel reduces the average reasoning length from 4,676 to 3,486 tokens while preserving the same accuracy. 
By dynamically reallocating penalties toward low-importance steps, \ourmodel selectively prunes redundant explorations while preserving the critical logical pivots necessary for correct solutions.
\section{Analysis}

\subsection{Performance under Different Budgets}

Figure~\ref{fig:dy_budget} presents the trade-off between Pass@1 accuracy and token usage on three mathematical reasoning benchmarks, comparing \ourmodel with state-of-the-art length-control methods.
Across all datasets and inference budgets, \ourmodel consistently achieves superior accuracy under comparable or smaller reasoning budgets.

On MATH-500, \ourmodel maintains strong performance even under aggressive compression.
Specifically, it achieves approximately 87.5\% accuracy with around 3.3k tokens, and still preserves over 86\% accuracy when the budget is reduced to roughly 1.6k tokens.
In contrast, baseline methods such as ThinkPrune and AdaptThink drop to around 84–85\% accuracy and exhibit pronounced degradation once the budget falls below 2.5k tokens.
This gap suggests that \ourmodel more effectively suppresses low-utility reasoning steps while preserving logical integrity.
% A similar trend is observed on AMC23.
% \ourmodel reaches over 80\% accuracy using approximately 3.2k tokens, whereas competing methods typically require 4k–5k tokens to approach comparable performance.
Under tighter budgets (below 3k tokens), baseline methods experience sharp accuracy declines, while \ourmodel remains substantially more robust.
On the more challenging AIME25 benchmark, \ourmodel continues to dominate the trade-off frontier.
At roughly 5.8k tokens, \ourmodel attains about 26.2\% accuracy, outperforming all baselines operating at similar or larger token budgets.
Even when reduced to approximately 4.6k tokens, \ourmodel maintains over 23\% accuracy, whereas other methods at comparable budgets remain below this level or require significantly longer responses.

\begin{figure}[ht]
    \centering
    \includegraphics[width=\linewidth]{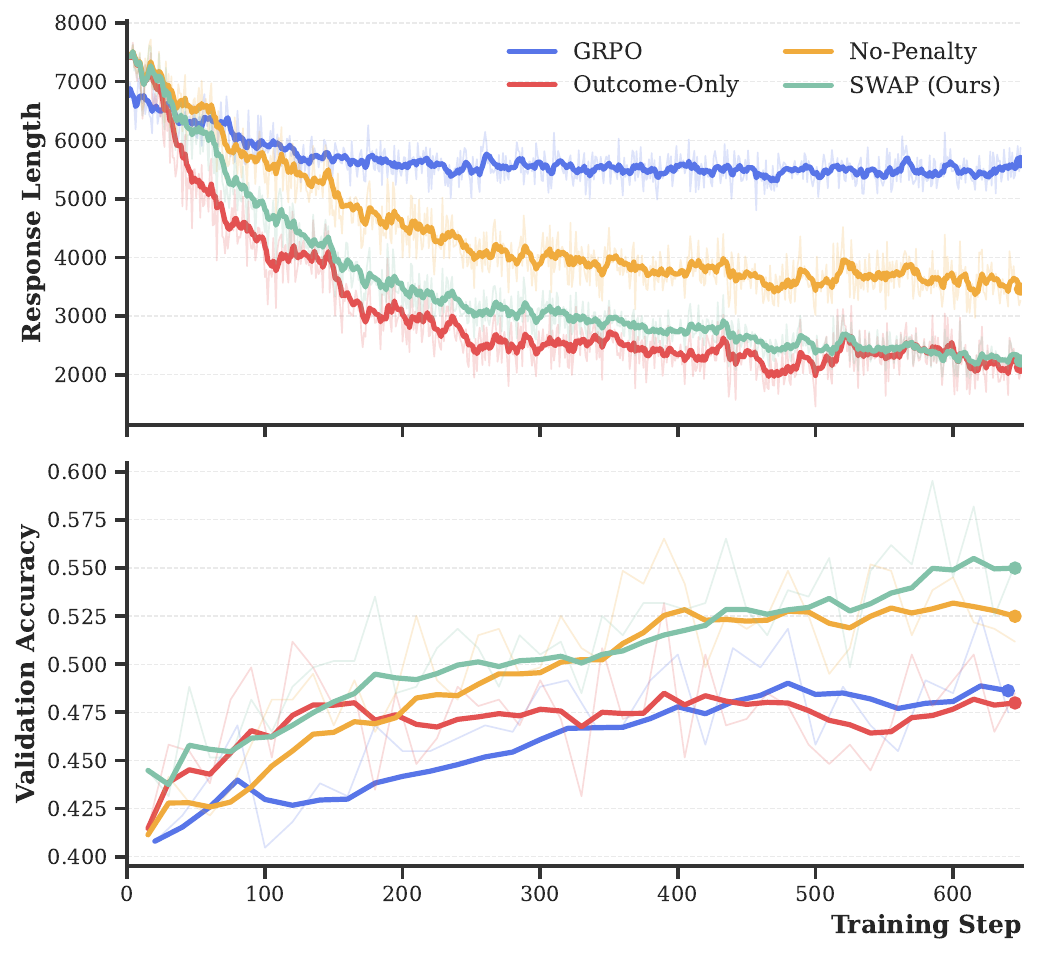}
    \caption{Training dynamics of different model ablations. \ourmodel achieves the best validation performance, with a final average response length similar to Outcome-Only (length penalty assigned at the trajectory level). }
    \label{fig:training_step}
    \vspace{-4mm}
\end{figure}

\subsection{Effect of Step Advantage Coefficient $\theta$}

We analyze the effect of the step advantage weight $\theta$ on accuracy and reasoning length in Figure~\ref{fig:theta}.
$\theta$ controls the strength of step-level process supervision, with larger values imposing stronger penalties on low-importance reasoning steps.
Across all benchmarks, increasing $\theta$ leads to a monotonic reduction in response length.
For small values of $\theta$ ($ \le 0.2$), \ourmodel preserves high accuracy while substantially reducing reasoning length, exhibiting a favorable efficiency–accuracy trade-off.
For example, on MATH-500, accuracy remains above 87.5\% as the average length decreases from 3{,}374 tokens at $\theta=0.05$ to 1{,}868 tokens at $\theta=0.2$.
% A similar trend is observed on AMC23, where \ourmodel maintains over 80\% accuracy while reducing response length by more than 40%.
As $\theta$ increases further, we observe a transition regime around $\theta \approx 0.4$–$0.5$, where accuracy begins to degrade more noticeably.
%On AMC23, accuracy drops from 78.3\% at $\theta=0.3$ to 71.5\% at $\theta=0.4$, while response length continues to decrease from 2{,}645 to 2{,}164 tokens.

This effect is more pronounced on harder benchmarks such as AIME24 and AIME25, where accuracy declines sharply beyond $\theta \approx 0.4$, indicating that overly strong step penalization starts to remove essential reasoning steps.
For large values of $\theta$ ($\theta \ge 0.7$), response length saturates across all datasets.
For instance, on AIME25, increasing $\theta$ from 0.7 to 1.0 yields only marginal length reduction (from 1{,}455 to 1{,}318 tokens), while accuracy remains low and relatively stable.
% This saturation suggests the existence of a minimum effective reasoning length per problem, beyond which further compression leads to limited gains and compromised correctness.
These results indicate that moderate values of $\theta$ (approximately 0.2–0.4) achieve the best balance between efficiency and accuracy.
In contrast, overly large $\theta$ leads to diminishing returns in length reduction and sharp accuracy degradation.
This behavior confirms that process advantages must be carefully weighted to suppress redundant reasoning without eliminating necessary intermediate supervision.

\subsection{Training Dynamics}

To study how different reward and penalty designs influence optimization, we analyze the training dynamics of response length and validation accuracy shown in Figure~\ref{fig:training_step}.
We compare standard GRPO, Outcome-Only (outcome reward with a global length penalty), No-Penalty (combined outcome and step rewards without length penalties), and \ourmodel.

Standard GRPO reduces response length slowly and converges to long reasoning traces (around 5{,}500 tokens), with relatively low validation accuracy.
Adding a global length penalty accelerates early length reduction; however, its accuracy remains noticeably lower, indicating that trajectory-level penalties alone encourage indiscriminate compression without improving reasoning quality.
In contrast, \ourmodel steadily reduces response length throughout training while simultaneously achieving the highest validation accuracy.
Compared to Outcome-Only and GRPO, our model converges to substantially shorter responses with better accuracy, and compared to No-Penalty, it achieves similar or higher accuracy with significantly fewer tokens.
These dynamics demonstrate that combining outcome grounding with progress-aware, step-weighted penalties enables both efficient compression and stable performance improvement during training.

\subsection{Additional Analysis}

We provide further analysis in Appendix~\ref{sec:appendix} to examine the robustness of our stepwise reward design. 
We study the sensitivity to step segmentation granularity in Figure~\ref{fig:step_size}. Extremely small step sizes lead to over-fragmentation and distorted importance estimation, while overly large step sizes weaken resolution and reduce length control. A moderate segmentation budget achieves the best accuracy–efficiency trade-off. 
In addition, Figure~\ref{fig:distribution} analyze the distribution of step-wise information gain before and after training. We observe a clear shift toward positive gain after optimization, indicating that low-utility reasoning steps are selectively suppressed while high-impact transitions are preserved. 
Detailed experimental results and additional discussions are provided in Appendix.

\section{Conclusion}

We presented \ourmodel, a step-wise length penalization framework that addresses overthinking in large reasoning models through fine-grained reinforcement learning.
By measuring each reasoning step’s intrinsic contribution via on-policy information gain and redistributing length penalties accordingly, \ourmodel enables selective compression that preserves critical reasoning while eliminating redundancy.
Integrated with a unified outcome–process advantage under GRPO, our approach balances correctness and efficiency without relying on external reward models or heuristic budgets.
Extensive experiments demonstrate that step-level optimization yields a substantially improved accuracy–efficiency trade-off across mathematical benchmarks.
Our findings suggest that overthinking is fundamentally a step-level phenomenon, and that stepwise credit assignment is a principled direction for efficient reasoning in future large-scale models.

\section*{Impact Statement}

This paper presents work whose goal is to advance the field of machine learning. There are many potential societal consequences of our work, none of which we feel must be specifically highlighted here.

\nocite{langley00}

\bibliography{icml2026}
\bibliographystyle{icml2026}

%%%%%%%%%%%%%%%%%%%%%%%%%%%%%%%%%%%%%%%%%%%%%%%%%%%%%%%%%%%%%%%%%%%%%%%%%%%%%%%
%%%%%%%%%%%%%%%%%%%%%%%%%%%%%%%%%%%%%%%%%%%%%%%%%%%%%%%%%%%%%%%%%%%%%%%%%%%%%%%
% APPENDIX
%%%%%%%%%%%%%%%%%%%%%%%%%%%%%%%%%%%%%%%%%%%%%%%%%%%%%%%%%%%%%%%%%%%%%%%%%%%%%%%
%%%%%%%%%%%%%%%%%%%%%%%%%%%%%%%%%%%%%%%%%%%%%%%%%%%%%%%%%%%%%%%%%%%%%%%%%%%%%%%
\newpage
\appendix
\onecolumn
\section{Appendix}
\label{sec:appendix}

\begin{figure*}[ht]
    \centering
    \includegraphics[width=\linewidth]{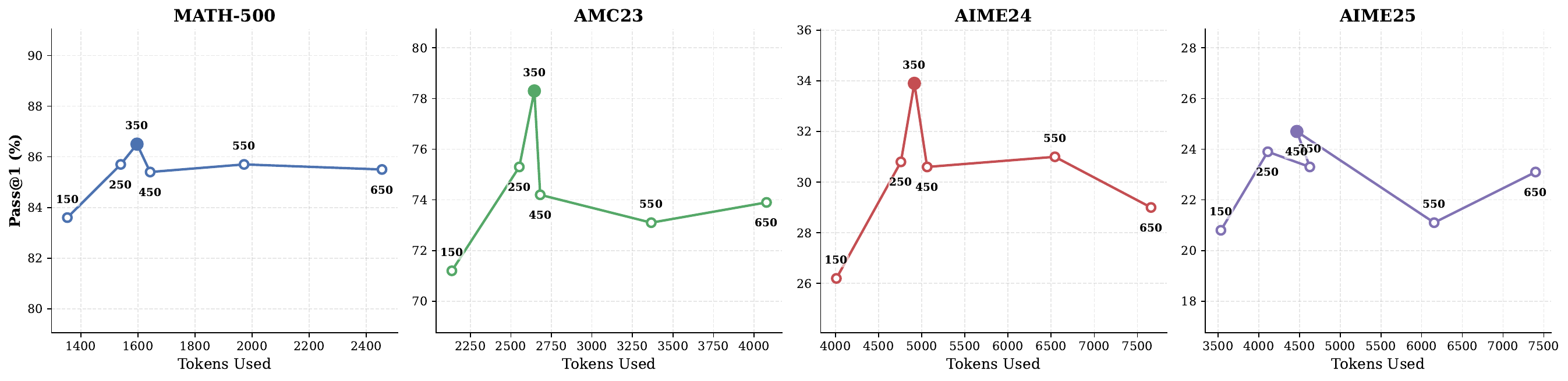}
    \caption{Comparison of Pass@1 accuracy vs. step length across four math benchmarks.}
    \label{fig:step_size}
\end{figure*}

\subsection{Effect of Step Size on Step Importance Estimation}
\label{sec:step_size}

We analyze the impact of step size $M$ in token-based step segmentation, which determines the granularity of step importance estimation.
Results are shown in Figure~\ref{fig:step_size}.
When the step size is too small, the response is split into many short steps, leading to excessive penalization of low-gain segments.
For example, with $M=150$, accuracy on MATH-500 drops to 83.6\% despite a short average length of 1{,}353 tokens, compared to 86.5\% at $M=350$.
Similar degradation is observed on AMC23 and AIME24, indicating that overly fine-grained steps distort step importance estimation.
Conversely, overly large step sizes reduce the resolution of step rewards and weaken length control.
On AIME24, increasing $M$ from 350 to 650 increases reasoning length from 4{,}915 to 7{,}658 tokens while accuracy decreases from 33.9\% to 29.0\%.
Across all benchmarks, $M=350$ achieves the best accuracy–efficiency trade-off, motivating our default choice.

\subsection{Distribution of Step-wise Information Gain}
\label{sec:distrubution}

\begin{figure*}[ht]
    \centering
    \includegraphics[
        width=\linewidth,
        trim=0.6cm 0.3cm 0.5cm 0.8cm,
        clip
    ]{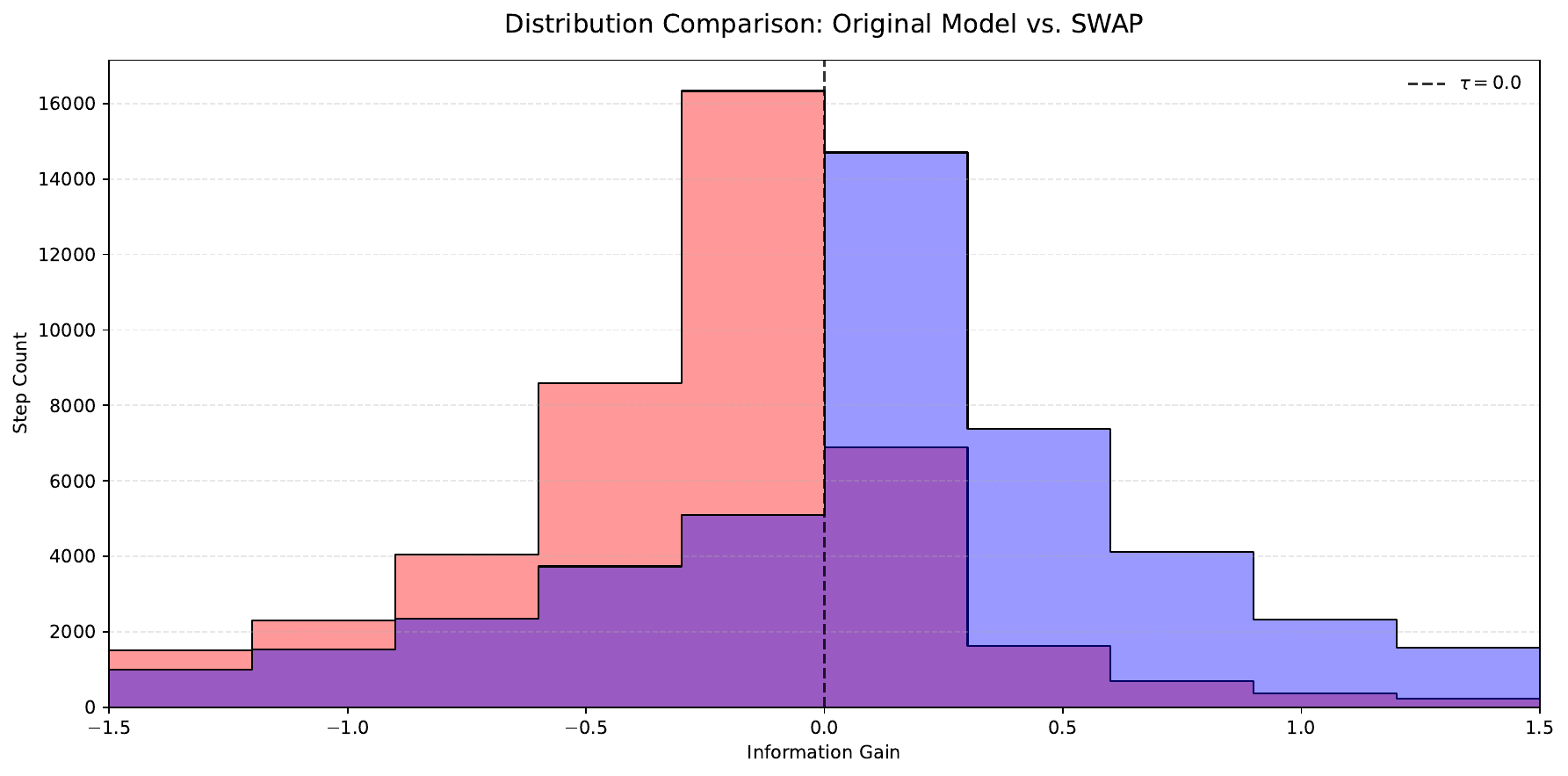}
    \caption{Distribution of step-wise information gain for the original model (red) and \ourmodel (blue), computed from 50k randomly sampled reasoning steps.
After training, \ourmodel exhibits a clear shift toward positive information gain, indicating that low-importance reasoning steps are suppressed while informative steps are preserved.}
    \label{fig:distribution}
\end{figure*}

Figure~\ref{fig:distribution} compares the distribution of step-wise information gain between the original model and \ourmodel, computed from 50k randomly sampled reasoning steps.
After training with stepwise reward penalization, the distribution shifts noticeably toward positive information gain.
In contrast to the original model, where a large fraction of steps exhibit near-zero or negative gain, \ourmodel concentrates mass on steps that contribute meaningful progress toward the final answer.
This indicates that \ourmodel effectively suppresses low-utility or redundant reasoning steps, preserving primarily high-impact reasoning transitions.

%%%%%%%%%%%%%%%%%%%%%%%%%%%%%%%%%%%%%%%%%%%%%%%%%%%%%%%%%%%%%%%%%%%%%%%%%%%%%%%
%%%%%%%%%%%%%%%%%%%%%%%%%%%%%%%%%%%%%%%%%%%%%%%%%%%%%%%%%%%%%%%%%%%%%%%%%%%%%%%

\end{document}